\documentclass[letterpaper]{article} 
\usepackage{aaai24}  
\usepackage{times}  
\usepackage{helvet}  
\usepackage{courier}  
\usepackage[hyphens]{url}  
\usepackage{graphicx} 
\usepackage{natbib}  
\usepackage{caption} 
\usepackage{algorithm2e}

\title{One step closer to unbiased aleatoric uncertainty estimation}
\author{
    Wang Zhang\textsuperscript{\rm 1},
    Ziwen Martin Ma\textsuperscript{\rm 2},
    Subhro Das\textsuperscript{\rm 3},\\
    Tsui-Wei Weng\textsuperscript{\rm 5},
    Alexandre Megretski\textsuperscript{\rm 1},
    Luca Daniel\textsuperscript{\rm 1},
    Lam M. Nguyen\textsuperscript{\rm 4}
}
\affiliations{

    \textsuperscript{\rm 1} Massachusetts Institute of Technology, Cambridge, MA, USA \\
    \textsuperscript{\rm 2} Harvard University, Cambridge, MA, USA \\
    \textsuperscript{\rm 3} MIT-IBM Watson AI Lab, IBM Research, Cambridge, MA, USA \\
    \textsuperscript{\rm 4} IBM Research, Thomas J. Watson Research Center, Yorktown Heights, NY, USA \\
    \textsuperscript{\rm 5} University of California, San Diego, CA, USA \\

}

\usepackage{bibentry}

\usepackage{amsmath}
\usepackage{amssymb}

\usepackage[capitalize,noabbrev]{cleveref}
\usepackage{graphics}
\usepackage{graphicx}
\usepackage{subfigure}
\usepackage{diagbox}

\usepackage{booktabs}  
\usepackage{makecell}
\usepackage{multirow}

\begin{document}

\maketitle

\begin{abstract}
Neural networks are powerful tools in various applications, and quantifying their uncertainty is crucial for reliable decision-making. In the deep learning field, the uncertainties are usually categorized into aleatoric (data) and epistemic (model) uncertainty. In this paper, we point out that the existing popular variance attenuation method highly overestimates aleatoric uncertainty. To address this issue, we propose a new estimation method by actively de-noising the observed data~\footnote{Source code available at \url{https://github.com/wz16/DVA}. Please refer to arXiv for full technical appendix.}. By conducting a broad range of experiments, we demonstrate that our proposed approach provides a much closer approximation to the actual data uncertainty than the standard method.
\end{abstract}

\section{Introduction}
Neural networks (NN) are capable of performing various regression and classification tasks in computer vision and natural language processing, ranging from basic multi-layer perceptron (MLP) that approximate straightforward functions to more sophisticated structures like transformers. As the capabilities of neural networks continue to expand, it becomes increasingly important to acknowledge the limitations of machine learning techniques. For instance, how does the corrupted data affect the trained model? And when does the model lack confidence in its predictions? Since most the neural network approaches have failed to capture the prediction confidence, quantifying the uncertainties for the deep learning methods becomes a crucial research topic.

In the machine learning literature, the sources of uncertainty can be classified into two categories \citep{der2009aleatory,hullermeier2021aleatoric,gal2016uncertaintythesis}: (1) aleatoric uncertainty, or data uncertainty, and (2) epistemic uncertainty, or model uncertainty. Aleatoric uncertainty is a measure of the inherent complexity or randomness that arises from the process of observing data from a true system, such as the noise in the measurements obtained from devices. This type of uncertainty is considered ``\textit{irreducible}" because no amount of additional data can eliminate the inherent stochasticity of the observation process. In contrast, epistemic uncertainty refers to the uncertainty that arises from the parameters of the model used to analyze the data, and can be reduced by increasing the size of the training data, which in turn increases the confidence in the model. Distinguishing between aleatoric and epistemic uncertainties can be beneficial in a wide range of applications. For example, aleatoric uncertainty can be used to recover the measurement tolerance of an instrument, while epistemic uncertainty can guide optimization algorithms during training \citep{choi2021active} or help to identify outliers during prediction \citep{seebock2019exploiting}.

There have been recent developments in deep learning research focused on techniques to capture aleatoric and epistemic uncertainties. To estimate the model uncertainty, one popular approach is to use Bayesian neural networks \citep{hernandez2015probabilistic,blundell2015weight}, which are neural networks that incorporate prior probability distributions over the model parameters. Another approach is to use ensemble methods \citep{parker2013ensemble,lakshminarayanan2017simple}, where multiple models are trained on the same dataset and their predictions are combined to produce a final prediction that incorporates epistemic uncertainties. \citet{gal2016dropout} showed the Monte Carlo dropout in deep learning approximates the Bayesian inference. A more recent branch directly estimates the aleatoric uncertainty together with the epistemic uncertainty \citep{lakshminarayanan2017simple,amini2020deep,valdenegro2022deeper}, by adding a parameterized variance term to the original stochastic prediction model. This addition aims to capture aleatoric uncertainty with a commonly used variance attenuation loss\footnote{see \Cref{eqn:loglikehoodatt}}. Although it may appear to be capable to handle two sources of uncertainty, this method has several significant drawbacks. \citet{seitzer2022pitfalls,stirn23Faithful} observed a decrease in the prediction performance when training it together with the variance head, which is attributed to the loss function being heavily biased on well-predicted samples with small variances. The prediction model performance difference presents additional challenges to the fair evaluation of the variance approximation module. Additionally, as will be further discussed in this paper\footnote{see \Cref{sec:VA_overestimate}}, the anticipated variance term under the ideal loss function exceeds solely the aleatoric uncertainty. Indeed, in the prior works on deep uncertainty estimation \citep{lakshminarayanan2017simple,amini2020deep,valdenegro2022deeper,seitzer2022pitfalls,stirn23Faithful}, the evaluation criteria are usually regression error or classification accuracy of the prediction model, rather than the quantitative difference between approximated uncertainty and true uncertainty. Therefore, gauging the effectiveness of uncertainty disentanglement becomes difficult.

The goal of this paper is to accurately identify the aleatoric uncertainty. We propose to split the prediction model training from the existing algorithm and develop a prediction-model-agnostic denoising approach that can better approximate the true noise level with a given trained prediction model. By augmenting a variance approximation module, this approach relies on the assumption that the distribution of the noise has zero mean, and seeks to remove the noise from the data in order to better capture the underlying signal. Doing so enables us to gain a more accurate understanding of the sources of uncertainty in the data, and leverage this information to improve our model and analysis.

\begin{figure*}[htb!]
\centering
\includegraphics[width=1.6\columnwidth]{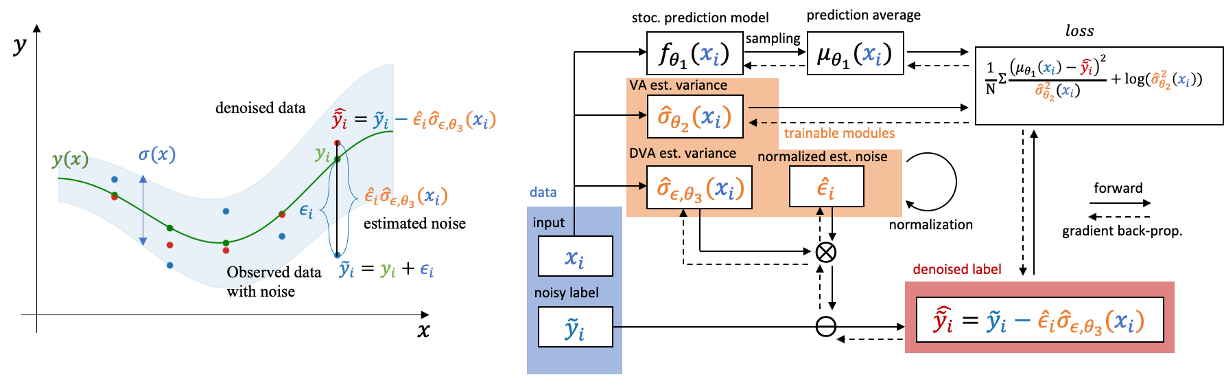}
\caption{Illustrative diagram and pipeline of the denoising variance attenuation method}
\label{fig:pipeline}
\end{figure*}

We summarize our main contribution as follows:
\begin{enumerate}
    \item Through theoretical analysis and experimental validation, we point out that a popular aleatoric uncertainty estimator does not identify the true data uncertainty. In fact, its expectation consists of aleatoric uncertainty and the squared expectation of the difference between the true labels and prediction average.
    \item Assuming the zero mean distribution of data noise, we propose a denoising method called Denoising Variance Attenuation (DVA), to actively infer the true data noise. We introduce additional parameterization on noise variance and normalized true noise, and optimize them through a customized projected gradient descent method. The method pipeline is illustrated in \Cref{fig:pipeline} with a detailed explanation in 
    \Cref{sec:proposedmethod}.
    \item From a theoretical standpoint, while our proposed method is still a biased estimator, it provides a more accurate approximation of the aleatoric uncertainty than the state-of-the-art alternatives. In practical applications, our approach yields estimates that are useful for capturing the true uncertainty of the data, and can be valuable to a wide range of problems.
\end{enumerate}


\section{Problem Formulation}
We start with a regression problem where the system input is $x\in \mathcal{X}\subset \mathbb{R}^n$ and output is $y\in \mathbb{R}^m$, where $\mathcal{X}$ is a compact set with positive volume. There exists a continuous mapping function $g(\cdot):\mathcal{X}\rightarrow \mathbb{R}^m$, which serves as the target function to regress. Consider a distribution over input feature $x$: $p(x)$, the finite dataset $D=\{(x_i,y_i)\}_{i=1}^M$ denotes $M$ sampled inputs from $p(x)$ and the corresponding outputs. When gathering data from the real world, it is common for the data to contain noise, which signifies the presence of aleatoric uncertainty. We denote the observed dataset by $\tilde{D}=\{(\tilde{x}_i,\tilde{y}_i)\}_{i=1}^M$, where $\tilde{x}_i=x_i+\xi_{i}, \tilde{y}_i=y_i+\epsilon_{i}$ are the noisy data and $\{\xi_{i},\epsilon_{i}\}$ are the i.i.d. realizations of the random noise variables $\xi, \epsilon$. In this paper, we assume the random noise distribution has zero-mean distribution and finite variance, i.e. $\mathbb{E}[\xi]=0,\mathbb{E}[\xi^2]=\sigma^2_\xi<\infty,\mathbb{E}[\epsilon]=0,\mathbb{E}[\epsilon^2]=\sigma^2_\epsilon<\infty$. When the noise variance is uniform across the system space, the aleatoric uncertainty is called ``\textit{homoscedastic}''; and, if the uncertainty varies against the system space, e.g. $\sigma^2_\xi(x),\sigma^2_\epsilon(x)$, the aleatoric uncertainty is considered ``\textit{heteroscedastic}''. In the classification problem, the output has the form of categorical labels, the aleatoric uncertainty arises from mislabeling or ambiguity between classes (e.g. the label is a probability distribution rather than a one-hot encoding).

Epistemic uncertainty evaluates the level of confidence in a model's prediction, which may be indirectly influenced by factors such as insufficient data, algorithmic randomness, or model structure choices. To capture the epistemic uncertainty, the model prediction is usually presented by a distribution rather than a deterministic value, e.g. $f_\theta(x)\sim N(\mu_\theta(x),s^2_\theta(x))$,  indicating that the model output follows a Gaussian distribution with mean and variance parameterized by $\theta$, and the variance identifies the epistemic uncertainty. 

Epistemic uncertainty relies on many factors (e.g., model choice, training algorithms, etc.), whereas aleatoric uncertainty is primarily an inherent property of the data. To assess the effectiveness of separating the two uncertainties, it is natural to compare the approximated data uncertainty with the true noise level. However, previous studies that claim to disentangle the two uncertainties \citep{lakshminarayanan2017simple,amini2020deep,seitzer2022pitfalls,valdenegro2022deeper,stirn23Faithful} often rely on the output regression error or classification accuracy as the benchmark criterion, rather than the direct comparison between the estimated aleatoric uncertainty and the true noise level. Our study aims to identify data noise (i.e., $\sigma^2_\xi,\sigma^2_\epsilon$) by quantitatively comparing the estimated noise levels with the true ones. As a byproduct, we develop a denoising strategy for noisy data, leading to improved prediction performance, evidenced by reduced regression error.

\subsection{Prior works on disentangling uncertainties}

Mean squared error (MSE) is commonly used as a loss function to train regression tasks, i.e. $\frac{1}{M}\Sigma_{i=1}^M (\tilde{y}_i-f_\theta(x_i))^2$, where $f_\theta(\cdot)$ is a continuous regression function parameterized by $\theta$ and can be either deterministic or stochastic. From a probabilistic viewpoint, minimizing MSE is equivalent to minimizing the negative log-likelihood of the observed data sampled from a Gaussian distribution centered at the prediction value, assuming a uniform noise variance $\hat{\sigma}^2$, 

\begin{align}
\label{eqn:loglikehooduniformsigma}
    & -log (\Pi_{i=1}^M  p(\tilde{y}_i|f_\theta(x_i))) \nonumber \\
    & = \frac{1}{M} \sum_{i=1}^M \frac{(\tilde{y}_i-f_\theta(x_i))^2}{2\hat{\sigma}^2} + \frac{log(\hat{\sigma}^2)}{2} + \textnormal{constant}.
\end{align}
Note that \Cref{eqn:loglikehooduniformsigma} differs from the MSE loss by a coefficient, the variance term $\hat{\sigma}^2$. In most optimization formulation, this term is usually neglected and therefore it fails to capture the uncertainty. When $f_\theta(\cdot)$ is stochastic (e.g. Bayesian Neural network), we let $f_{\theta}(\cdot)=\mu_{\theta}(\cdot)+\epsilon_{\theta}(\cdot)$, where $\mu_{\theta}(\cdot)$ and $\epsilon_{\theta}(\cdot)$ denote the deterministic mean and epistemic random component with a variance of $s^2_{\theta}(\cdot)$. Inspired by \citet{nix1994variance}, a popular method \citep{lakshminarayanan2017simple,amini2020deep,valdenegro2022deeper} replaces the fixed variance with a predicted one, to identify the aleatoric uncertainty. This modified loss function is called variance attenuation (VA):

\begin{equation}\label{eqn:loglikehoodatt}
    L_{va} = \frac{1}{M}  \sum_{i=1}^M \frac{(\tilde{y}_i-\mu_{\theta_1}(x_i))^2}{2\hat{\sigma}^2_{\theta_2}(x_i)} + \frac{log(\hat{\sigma}^2_{\theta_2}(x_i))}{2},
\end{equation}
where $\hat{\sigma}_{\theta_2}(\cdot)>0$ is a parameterized and continuous estimator for aleatoric uncertainty, and $\mu_{\theta_1}(\cdot)$ is the mean of stochastic prediction $f_{\theta_1}(\cdot)$. For notation clarity, we rewrite the parameter of the regression function and variance estimator as $\theta_1$ and $\theta_2$, respectively.

Note that the above estimator implicitly assumes the input feature $x$ has zero noise ($\sigma^2_\xi(x)\equiv 0$). Besides, the additional variance head in \Cref{eqn:loglikehoodatt} does not affect the original model $f_{\theta_1}(\cdot)$, the epistemic uncertainty is still incorporated in the distribution of $f_{\theta_1}(\cdot)$. There exist many quantification approaches for epistemic uncertainty. In Bayesian neural networks \citep{blundell2015weight}, a posterior distribution of the neural network parameter $\theta$ is inferred during training, and given a testing input, the prediction distribution can be approximated by drawing samples of ${\theta}$. For ensemble methods, multiple samplings are also required to approximate $f_{\theta_1}(\cdot)$. \citet{amini2020deep} leverages Normal Inverse-Gamma (NIG) distribution and proposes Deep Evidential Regression (DER) to analytically output the mean and variance of the prediction distribution along with the aleatoric uncertainty. This approach streamlines the sampling process and enables efficient uncertainty learning.

Once $\mu_{\theta_1}(\cdot),s^2_{\theta_1}(\cdot)$ are obtained from the prediction model, $\mu_{\theta_1}(\cdot)$ is incorporated into \Cref{eqn:loglikehoodatt} to help model to estimate $\hat{\sigma}^2_{\theta_2}(\cdot)$. This completes the framework of disentangling uncertainties, where $s^2_{\theta_1}(\cdot)$ and $\hat{\sigma}^2_{\theta_2}(\cdot)$ handles epistemic and aleatoric uncertainty respectively.

\subsection{Limitations of aleatoric estimator}\label{sec:VA_overestimate}

In this section, we argue that the aleatoric estimator $\sigma_{\theta_2}(\cdot)$ in \Cref{eqn:loglikehoodatt} is imperfect and does not capture the exact aleatoric uncertainty. Consider a homoscedastic problem with uniform label noise variance $\mathbb{E}[(\tilde{y}-y)^2]=\sigma_\epsilon^2$, and $\hat{\sigma}^2_{\theta_2}$ being a single optimizable scalar parameter rather than a function. Fix the regression function parameter $\theta_1$ and optimize \Cref{eqn:loglikehoodatt} over $\hat{\sigma}^2_{\theta_2}$, where the optimization parameter is a local minimum. Then, it reduces to,

\begin{align}\label{eqn:flawedestimatorexpecthomo}
    & \frac{dL_{va}}{d\hat{\sigma}^2_{\theta_2}} = \sum_{i=1}^M  \frac{-(\tilde{y}_i-\mu_{\theta_1}(x_i))^2}{2\hat{\sigma}^4_{\theta_2}} + \frac{1}{2\hat{\sigma}^2_{\theta_2}} =0,\nonumber \\
    \implies & \hat{\sigma}^2_{\theta_2}  = \frac{1}{M}\sum_{i=1}^M  (\tilde{y}_i-\mu_{\theta_1}(x_i))^2 ,\nonumber \\
    \implies & \mathbb{E}[\hat{\sigma}^2_{\theta_2}] = \mathbb{E}_{x_i,\epsilon_i}[ (y_i+\epsilon_i-\mu_{\theta_1}(x_i))^2] 
    \nonumber \\
    & \quad \quad \quad = \mathbb{E}[ (y-\mu_{\theta_1}(x))^2] +\sigma_\epsilon^2.
\end{align}

where, $\sigma^2_{\epsilon}$ is the variance of the data noise, $\mu_{\theta_1}(x)$ is the mean of the epistemic regression model. \Cref{eqn:flawedestimatorexpecthomo} says that when the loss function reaches optimal, the estimator $\hat{\sigma}^2_{\theta_2}(\cdot)$ overestimates the aleatoric uncertainty by an expectation of square difference between the clean data and mean of regression predictions. The above derivation is extendable to a heteroscedastic setup by parameterizing $\hat{\sigma}^2_{\theta_2}(\cdot)$ to a function of $x$ and analyzing the necessary condition for the extreme points. The conclusions are similar for the heteroscedastic cases and the analysis is deferred to the \Cref{appendix:theory}.

\section{Proposed Method}
\label{sec:proposedmethod}
In the previous section, we presented a theoretical argument highlighting the substantial bias present in the aleatoric uncertainty estimates obtained via the popular VA method. Here, we introduce a novel method that is developed to mitigate this bias and to offer more precise estimates\footnote{Our method incorporates an additional module to improve variance estimation, rather than retrain the regression model $f_{\theta_1}(\cdot)$ from scratch.}.

\subsection{Normalized gradient descent denoising}
When dealing with a dataset of a sufficiently large size, an intriguing opportunity arises to take advantage of the property that the mean value of noise realizations tends toward zero. By leveraging this characteristic of large datasets, we can design algorithms that recover individual noise values for each data point and the corresponding noise variance, providing a more precise estimate of the noise levels present in the data.

Assuming that the examples have only label noise, for each data point $(x_i,\Tilde{y}_i)$ where $\Tilde{y}_i=y_i+\epsilon_i$, we augment a normalized estimated noise $\hat{\epsilon}_i$. For the noise variance, we employ an estimating function $\hat{\sigma}^2_{\epsilon}(\cdot)$ which can be either heteroscedastic or homoscedastic. We aim to approximate the true noise $\epsilon_i$ by $\hat{\epsilon}_i\hat{\sigma}_{\epsilon}(x_i)$, or estimate the clean data $y_i$ by denoising the observed data $\Tilde{y}_i-\hat{\epsilon}_i\hat{\sigma}_{\epsilon}(x_i)$.

In this paper, we propose a normalized gradient descent type algorithm, named Denoising Variance Attenuation (DVA), which is inspired by the Projected Gradient Descent (PGD) Attack~\citep{madry2017towards,kurakin2018adversarial}. The PGD Attack is a technique used to generate robust adversarial examples by iteratively updating strong adversaries using the loss function gradient and projecting them into the $\mathcal{L}_p$ neighborhood of the initial input. Our proposed method applies a similar process to denoise inputs or outputs, by optimizing over the estimated noise and increasing the log-likelihood of the denoised data while ensuring that the total of normalized noises has zero mean and unit variance. The illustrative diagram and pipeline of the DVA method are shown in \Cref{fig:pipeline}. Let the estimated variance function be parameterized by $\hat{\sigma}_{\epsilon,\theta_3}(\cdot)$, and we formulate the constrained optimization problem on the denoising variance attenuation loss:

\begin{align}\label{eqn:denoisevarianceatt}
    & \min_{\theta_2,\theta_3,\{\hat{\epsilon}_i\}_{i=1}^M}  L_{dva}, \text{ where} \nonumber  \\L_{dva} &=\frac{1}{M} \sum_{i=1}^M \frac{(\tilde{y}_i-\hat{\epsilon}_i\hat{\sigma}_{\epsilon,\theta_3}(x_i)-\mu_{\theta_1}(x_i))^2}{2\hat{\sigma}^2_{\theta_2}(x_i)} + \frac{log(\hat{\sigma}^2_{\theta_2}(x_i))}{2}, \nonumber \\
    & s.t. \quad \sum_{i=1}^M \hat{\epsilon}_i/M = 0,  \sum_{i=1}^M \hat{\epsilon}_i^2/M=1.
\end{align}

We implement the constrained optimization in a straightforward way. After each gradient update, we perform a normalization procedure for all the inferred noise, to ensure that the estimated normalized noises have zero mean and unit variance. The key steps of the algorithm are listed below, with the detailed algorithm being deferred to \Cref{appendix:algo}.
\begin{align}
     \text{Gradient step: }&\theta_2 \leftarrow \theta_2-\eta \frac{dL_{dva}}{d\theta_2};\theta_3 \leftarrow \theta_3-\eta \frac{dL_{dva}}{d\theta_3};\nonumber\\
    & \hat{\epsilon}_i \leftarrow \hat{\epsilon}_i-\eta \frac{dL_{dva}}{d\hat{\epsilon}_i}, \text{ for } i=1,2...M; \nonumber
\end{align}
\begin{align}\label{eqn:pseudoDVAalg}
    \text{Normalization step: } &\mu \leftarrow \sum_{i=1}^M \hat{\epsilon}_i/M; \nonumber\\
     &var \leftarrow \sum_{i=1}^M (\hat{\epsilon}_i-\mu)^2/M;\nonumber\\
    & \hat{\epsilon}_i \leftarrow \frac{\hat{\epsilon}_i-\mu}{\sqrt{var}}, \text{ for } i=1,2...M.  
\end{align}

Note that this optimization process is not dependent on the training of the regression model, $f_{\theta_1}(\cdot)$. In practice, they can be trained simultaneously or separately.

\subsection{Necessary conditions for constrained optimization}

We start with a uniform noise variance and assign a single optimizable parameter for both $\hat{\sigma}^2_{\theta_2}$ and $\hat{\sigma}^2_{\epsilon,\theta_3}$. Consider the problem in \Cref{eqn:denoisevarianceatt}, when the optimization parameter is a local minimum, we have the following necessary conditions for constrained optimization with multiplier $\lambda_1,\lambda_2$:

\begin{align}\label{eqn:kktcondition}
     & \frac{-2(\Tilde{y}_i-\hat{\epsilon}_i\hat{\sigma}_{\epsilon,\theta_3} -\mu_{\theta_1}(x_i))\hat{\sigma}_{\epsilon,\theta_3}}{2\hat{\sigma}^2_{\theta_2}}+\lambda_1+2\lambda_2 \hat{\epsilon}_i =0, \nonumber \\
    & \text{ for } i = 1,2...M; \nonumber \\
    & \sum_{i=1}^M \frac{-2(\Tilde{y}_i-\hat{\epsilon}_i\hat{\sigma}_{\epsilon,\theta_3} -\mu_{\theta_1}(x_i))\hat{\epsilon}_i}{2\hat{\sigma}^2_{\theta_2}}=0, \quad\nonumber  \\
       & \sum_{i=1}^M \frac{(\Tilde{y}_i-\hat{\epsilon}_i\hat{\sigma}_{\epsilon,\theta_3} -\mu_{\theta_1}(x_i))^2}{M} = \hat{\sigma}^2_{\theta_2}.
\end{align}
While the derivation of this constrained optimization is deferred to \Cref{appendix:KKTderivation}, its solution yields,
\begin{equation*}
    \hat{\sigma}_{\epsilon,\theta_3}^2 = \frac{1}{M}\sum_{i=1}^M (\Tilde{y}_i -\mu_{\theta_1}(x_i))^2-\frac{1}{M^2} \left[ \sum_{i=1}^M (\Tilde{y}_i -\mu_{\theta_1}(x_i)) \right]^2
\end{equation*}
    
\begin{align}\label{eqn:kktresults}
    \mathbb{E}[\hat{\sigma}_{\epsilon,\theta_3}^2] & =\frac{M-1}{M} \sigma_\epsilon^2+\frac{M-1}{M}\mathbb{E}[(y-\mu_{\theta_1}(x))^2]\nonumber\\
    &-\frac{M-1}{M}\mathbb{E}^2[y-\mu_{\theta_1}(x)].
\end{align}

With $M\rightarrow\infty$, $\frac{M-1}{M} \rightarrow 1$,
\begin{align*}
    \mathbb{E}[\hat{\sigma}_{\epsilon,\theta_3}^2] & \rightarrow \sigma_\epsilon^2 +\mathbb{E}[(y-\mu_{\theta_1}(x))^2]- \mathbb{E}^2[y-\mu_{\theta_1}(x)]\nonumber\\
    & = \sigma_\epsilon^2 + Var[y-\mu_{\theta_1}(x)].
\end{align*}


Our new estimator still overestimates the aleatoric uncertainty by $var[y-\mu_{\theta_1}(x)]$; however, it is smaller by a term of $\mathbb{E}^2[y-\mu_{\theta_1}(x)]$ compared to the expectation of the previous estimator in \Cref{eqn:flawedestimatorexpecthomo}. This brings us one step closer to capturing aleatoric uncertainty. Further theoretical analysis on other setups is provided in \Cref{appendix:theory}.

\subsection{Segmented normalization for heteroscedastic setting}

In estimating the true realization of the noise, we rely on the law of large numbers and unbiased estimation of the variance theorem  \citep{degroot2012probability} that the mean and variance of a large number of i.i.d. samples converge to the expected value of the underlying distribution. However, it should be noted that the normalization process in \Cref{eqn:pseudoDVAalg} only ensures that the noise has zero mean and unit variance, without enforcing the independent and identically distributed property. In practice, if the normalization is performed globally, it is possible to encounter situations where the variances of the noise are dependent on the location (e.g. input $x$) and such imbalances are corrected by the variance estimator $\hat{\sigma}_{\epsilon,\theta_3}(x)$ leading to its inaccurate estimation. To eliminate such an effect, we propose segmented normalization (with details in \Cref{appendix:algo}), where we segment the whole training dataset into multiple components and perform normalization individually to prevent the dependence on the local variance.
\section{Experiments}

To showcase and validate the effectiveness of the DVA technique, we start with a simple regression example. This example helps to investigate the effects of various noise types, such as input and label noise, as well as heteroscedastic and homoscedastic noise. Our approach exhibits strong performance when handling label noise. Detecting input noise typically poses a notable challenge due to the non-linear mapping of the prediction model. Nevertheless, we successfully apply our approach to a problem involving the identification of a dynamic system, in which both input and label are influenced by the same source of noise. To further demonstrate the applicability of our method, we then apply it to real-world tasks involving age prediction and depth estimation. The outcomes underscore the effectiveness of our strategy, even when dealing with large-scale problems.

\subsection{Toy regression example}
We begin with a simplified regression scenario where the underlying mapping function is $f(x)=x(1+sin(x))$. We examine various noise configurations and compare our denoising variance attenuation (DVA) technique with the traditional variance attenuation (VA) method. We investigate two different noise configurations:
\begin{itemize}
    \item Homoscedastic label noise: $\sigma_{\xi}(x)=0,\sigma_{\epsilon}(x)=a$,
    \item Heteroscedastic label noise: $\sigma_{\xi}(x)=0,\sigma_{\epsilon}(x)=a(1+0.1x)$,
\end{itemize}
where $a>0$ is a constant that represents the magnitude of the standard deviation of the noise. Although we do not make any assumptions on the noise distribution in our analysis, for this example we generate the noise by sampling from a Gaussian distribution with a specified variance (e.g. $N(0,\sigma^2_{\epsilon}(x))$).

The parameterization of both VA variance ($\hat{\sigma}^2_{\theta_2}$) and DVA variance ($\hat{\sigma}_{\epsilon,\theta_3}^2$) follow the noise setup. If the noise is homoscedastic, we use a single scalar parameter with global normalization of noises, and if the noise is heteroscedastic, we use a neural network to parameterize the variance dependence on $x$ and perform local segmented normalization with 10 segments.  

To capture the epistemic uncertainty, we employ two stochastic prediction models: an ensemble model with 5 base learners and Bayesian neural network (BNN, \citealt{hernandez2015probabilistic}) with a sampling size of 5. For a fair comparison between the VA and DVA, the prediction model is pre-trained with MSE loss ahead of the variance training. A total of 1000 samples are drawn from $x\in[1,9]$ for training. Each experiment is performed on 5 random seeds for reproducibility. For more details on the experimental setup, please refer to \Cref{appendix:experimentdetails}.

In the homoscedastic problem, we increase the global label noise variance from $0.5$ to $8.0$, and evaluate the performance across different experimental setups. The results, as shown in \Cref{tab:xplusonebysinx_homo_y}, indicate that our method still tends to overestimate the true variance. However, the gap between the estimated and true values is considerably smaller than that of the VA method across all experiment setups.

\begin{table}[htbp!]
\centering
\caption{Aleatoric uncertainty estimation under homoscedastic label noise (closer to $a^2$ is better)}\label{tab:xplusonebysinx_homo_y}
\scalebox{0.8}{
\begin{tabular}{|l|*{5}{c|}}
        \hline
        Method&$a^2=$0.5&1.0&2.0&8.0\\\hline
        VA + BNN & 2.41$\pm$0.20 &3.20$\pm$0.36 & 4.02$\pm$0.26  
        & 10.35$\pm$0.32 \\\hline
        DVA + BNN & \textbf{1.12$\pm$0.20}  &  \textbf{1.87$\pm$0.25}  & \textbf{2.54$\pm$0.31} 
        & \textbf{9.03$\pm$0.65} \\\hline
        VA + ensemble & 2.53$\pm$ 1.21 & 2.60$\pm$0.68 & 4.72$\pm$0.41  
        &10.54$\pm$1.09 \\\hline
        DVA + ensemble&  \textbf{1.77$\pm$0.81} & \textbf{2.28$\pm$0.61}   & \textbf{3.48$\pm$0.49}  & 
        \textbf{9.34$\pm$0.30} \\\hline
\end{tabular}}
\end{table}

\begin{figure*}[htb!]
\centering
\includegraphics[width=1.4\columnwidth]{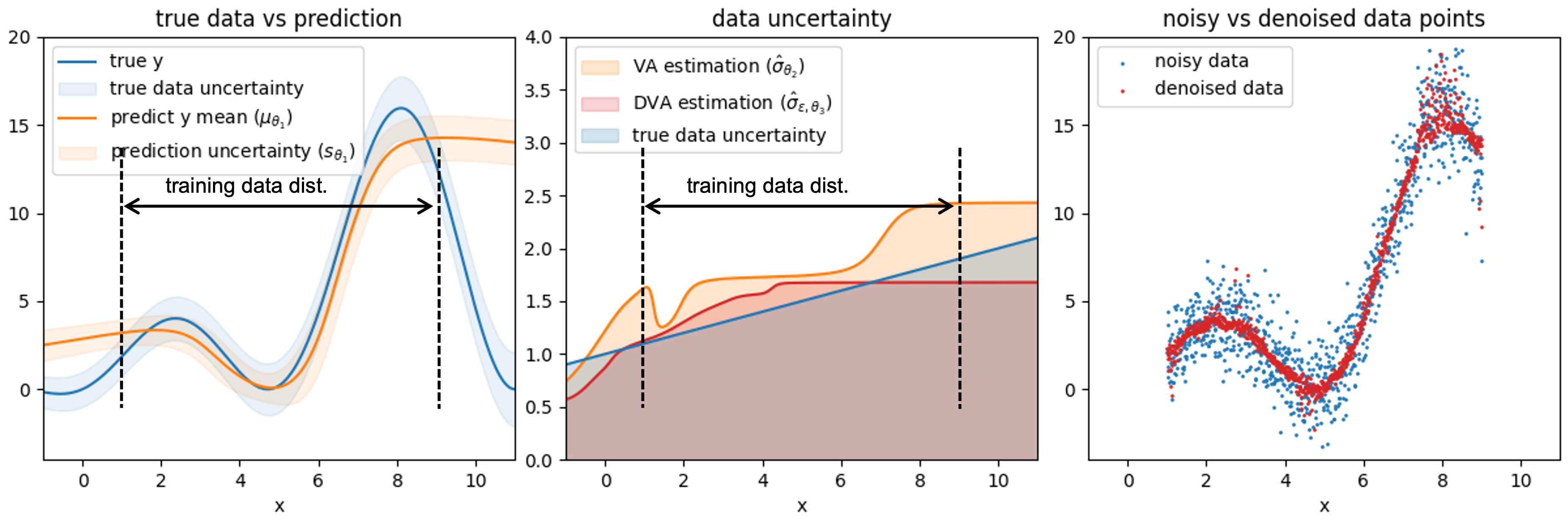}
\caption{Visualization of the toy example experiment under heteroscedastic label noise}
\label{fig:heterexperiment}
\end{figure*}

In the heteroscedastic problem, we assign a $x\text{-}$dependent noise variance $\sigma_\epsilon(x)=a(1+0.1x)$ and use squared mean error between the estimation and the true variance as a comparison criterion. Our proposed DVA estimation is closer to the true label variance across almost all experiment setups. The results of a typical experiment when $a=1$ are shown in \Cref{fig:heterexperiment}. The left figure displays the data and model uncertainties, the middle figure visualizes the estimated data uncertainty, and the right figure demonstrates the comparison between the noisy data observations and their corresponding denoised counterparts. Specifically, the VA method tends to overestimate the data uncertainty across the whole input domain, while the DVA method provides a more accurate estimation. The DVA method exhibits a flat line around $x=8$, this corresponds to the inferior denoising performance in that specific region, which is attributed to the bias in the prediction model.

\begin{table}[htbp!]
\centering
\caption{Aleatoric uncertainty estimation under heteroscedastic label noise, squared mean difference to true variance (smaller is better)}\label{tab:xplusonebysinx_heter_y}
\scalebox{0.8}{
\begin{tabular}{|l|*{5}{c|}}
        \hline
        Method&$a^2=$0.5&1.0&2.0&8.0\\\hline
        VA + BNN & 0.65$\pm$0.17 &\textbf{0.32$\pm$0.06} & 0.36$\pm$0.17 
        & 0.51$\pm$0.18 \\\hline
        DVA + BNN & \textbf{0.45$\pm$0.26}  &  0.40$\pm$0.61  & \textbf{0.32$\pm$0.17}  
        & \textbf{0.37$\pm$0.17} \\\hline
        VA + ensemble & 1.97$\pm$ 0.86 & 1.30$\pm$ 0.44 & 1.97$\pm$0.24  
        &1.32$\pm$0.32 \\\hline
        DVA + ensemble&  \textbf{1.42$\pm$1.17} & \textbf{0.82$\pm$0.42}   & \textbf{1.00$\pm$0.28}  
        &\textbf{1.23$\pm$0.39} \\\hline

\end{tabular}}
\end{table}

In addition to the label noise cases, we also perform experiments on scenarios with input noise ($\sigma_{\xi}(x)=a,\sigma_{\epsilon}(x)=0$) and defer the results to \Cref{appendix:inputnoise}. In general, we observe a trend of increasing noise, but the estimated noise tends to be exaggerated. This is due to the denoised data inferred from neural networks starting to converge towards regions outside the known distribution, as the networks' ability to accurately predict beyond existing data diminishes.

\subsection{Dynamical system noise quantification with neural ODE identification}

For general regression problems with both input and output noise, our method is not effective because it cannot distinguish the weights of the noise sources \footnote{Consider an identity mapping problem where $y=x$ with  $\sigma_{\xi}^2=\sigma_{\epsilon}^2=2$. If we apply our denoising method on both input and labels, we may end up with any arbitrary combination of $(\hat{\sigma}_{\xi}^2,\hat{\sigma}_{\epsilon}^2)$ that satisfies $\hat{\sigma}_{\xi}^2+\hat{\sigma}_{\epsilon}^2=4$).}. However, for dynamical systems which take initial conditions as inputs and trajectory observations as labels, the inputs and labels are subject to the same sources of inherent uncertainty. As a result, we can effectively model this uncertainty using the same set of parameters. We conducted experiments with a synthetic dynamical system ($\dot{x}=x(1+sin(x))$) and sampled 100 trajectories with 51 points/trajectory and homoscedastic Gaussian noise as training data. We learn the dynamical system by employing neural ODE \citep{chen2018neuralode} with Bayesian parameters and estimate the uncertainty by direct MSE loss and DVA method. \Cref{tab:neuralode_xplusonebysinx_homox} shows the DVA estimation is much closer than the MSE estimation.

\begin{table}[htbp!]
\centering
\caption{Dynamical system aleatoric uncertainty estimation under homoscedastic label noise (closer to $a^2$ is better)}\label{tab:neuralode_xplusonebysinx_homox}
\scalebox{0.8}{
\begin{tabular}{|l|*{5}{c|}}
        \hline
        Method&$a^2=$0.5&1.0&2.0&8.0\\\hline
        MSE est. + BNN & 1.87$\pm$0.56 &2.69$\pm$0.87 & 4.00$\pm$1.73 
        & 11.63$\pm$4.82 \\\hline
        DVA + BNN & \textbf{1.06$\pm$0.04}  &  \textbf{1.66$\pm$0.10}  & \textbf{2.86$\pm$0.19} 
        & \textbf{10.55$\pm$0.45} \\\hline
\end{tabular}}
\end{table}

\vspace{-0.2in}
\subsection{Depth estimation}

We demonstrate the applicability of our DVA method to a depth estimation task. The NYU Depth v2 dataset \citep{silberman2012indoor} contains 27k RGB-depth image pairs. These images represent indoor scenes such as kitchens, and the pixel labels share the same dimensions as the input. We utilize the VPD (Visual Perception with a pre-trained Diffusion model, \cite{zhao2023unleashing}) model trained on the dataset as a prediction model.  We add different levels of homoscedastic synthetic Gaussian noise to the label pixels and apply our DVA method to detect the noise level. The proposed DVA method outperforms the VA method in better estimating the aleatoric uncertainty.

\begin{figure}[htb!]
\centering
\includegraphics[width=0.8\columnwidth]{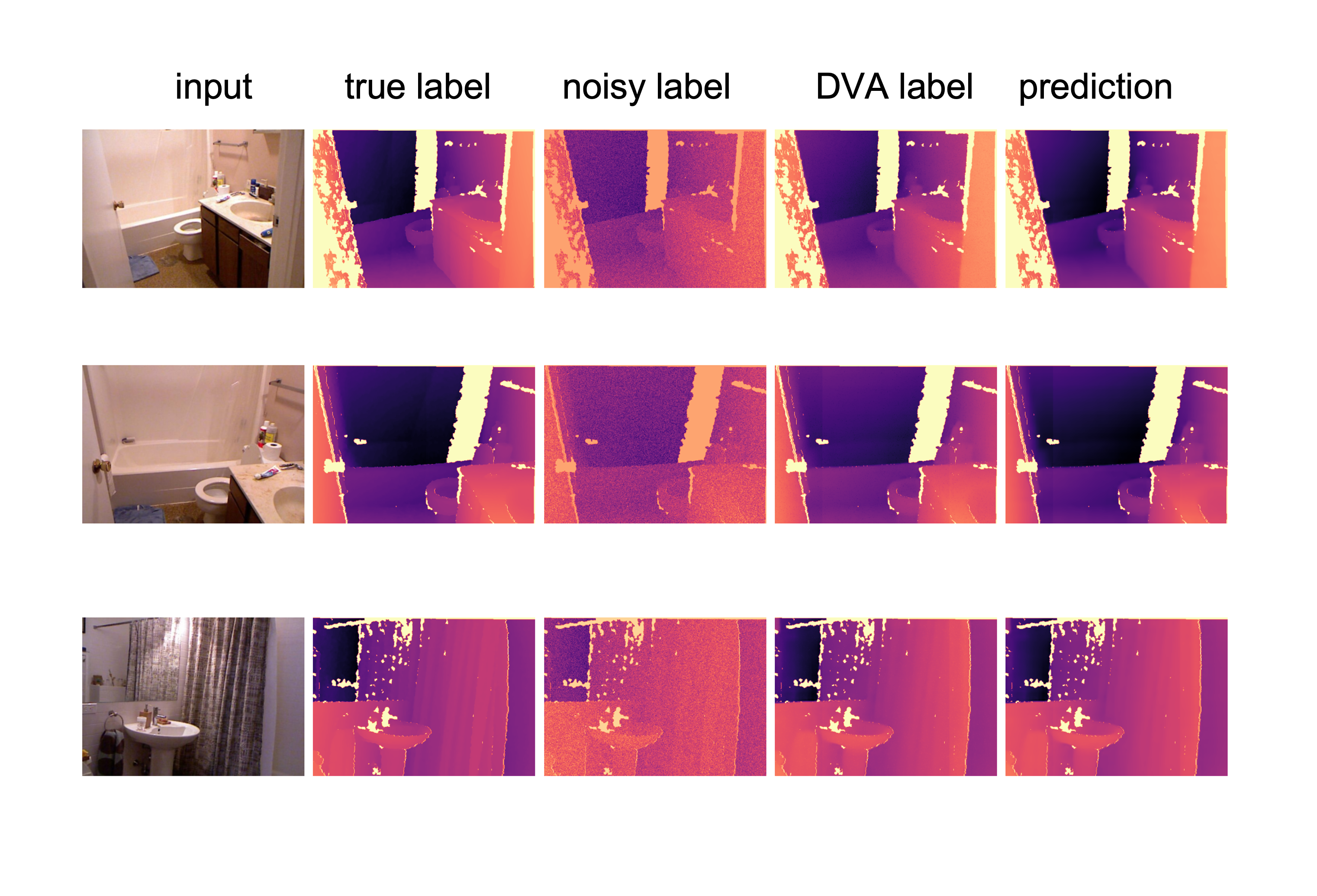}
\caption{Sample examples under $a^2=0.25$}
\label{fig:depthprediction}
\end{figure}

\begin{table}[htb!]
\centering
\caption{Aleatoric uncertainty estimation under homoscedastic input noise in depth estimation}\label{tab:depth_homo}
\scalebox{0.8}{
\begin{tabular}{|l|*{4}{c|}}
        \hline
        Method&$(a^2)=$0.01&0.25&1.0\\\hline
        VA + VPD  & $0.105 \pm 0.001$ & $0.344 \pm 0.001$ & $1.093 \pm 0.001$ \\\hline
        DVA + VPD & \textbf{0.021 $\pm$ 0.000} & \textbf{0.281 $\pm$ 0.001} & \textbf{1.056 $\pm$ 0.006} \\\hline
\end{tabular}}
\end{table}

\subsection{Age prediction}
\begin{figure}[htb!]
\centering
\includegraphics[width=1\columnwidth]{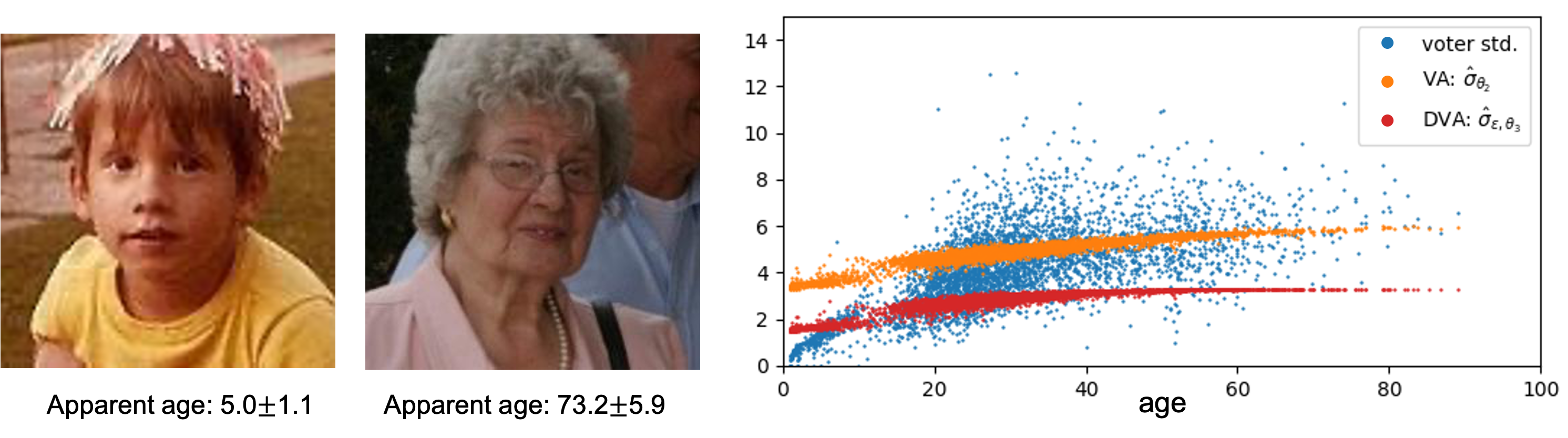}
\caption{Left: samples, Right: apparent age std. (from voters) vs estimated data uncertainty}
\label{fig:ageprediction}
\end{figure}

We further extend our DVA method to real-world data by performing an age prediction task by utilizing the APPA-REAL database \citep{agustsson2017apparent} (\Cref{fig:ageprediction} shows two samples) consisting of 7591 images. Each image in the database has been assigned an apparent age label based on an average of 38 individual votes per image. We employ a ResNext-50 (32$\times$4d) pretrained on ImageNet, replace the last layer and fine-tune it on the APPA-REAL database with averaged apparent age as labels. 

To estimate uncertainty, we append an additional linear layer to the second last layer of the network, which maps its outputs (with a dimension of 2048) to a logarithmic uncertainty measure. We then fine-tune only this linear layer. In \Cref{fig:ageprediction}, we present the VA and DVA estimates of the uncertainty associated with the averaged apparent age, as well as the standard deviation of the voters' responses for the entire training dataset. Viewing the voters as prediction models that receive images without necessarily reflecting the true underlying age, their forecasted variability encompasses two components: aleatoric uncertainty (discrepancy between the image-based age and the image itself) and epistemic uncertainty (disparity between image-based age and the individual predicted age by the voter). It is worth noting that the epistemic gap between individual voting and group-averaged voting has already been eliminated due to the averaging of labels. Consequently, the aleatoric uncertainty of the averaged labels should be lower than the standard deviation of the voting results indicated by the blue dots in the figure. Our proposed DVA method better approximates this lower bound, particularly for children (whose ages are generally easier to determine accurately than that of adults).

\section{Discussion}
\subsection{Significance of aleatoric uncertainty quantification}
Aleatoric uncertainty analysis stands as a pivotal facet across a multitude of scientific and engineering domains, contributing significantly to comprehending the inherent fluctuations within a system or the observation process. Through quantifying aleatoric uncertainty, researchers attain a deeper awareness of the data's intrinsic randomness. This knowledge subsequently empowers researchers to construct more resilient models capable of accommodating and flexibly responding to conditions with built-in uncertainty. This becomes particularly pronounced in the realm of forecasting models, such as dynamical systems, which heavily rely on observational measurements as input. The ability to precisely quantify measurement uncertainty equips researchers with the means to not only evaluate the trustworthiness of their data but also to effectively incorporate the appropriate level of variability into their forecasts. In essence, aleatoric uncertainty analysis serves as an essential part when dealing with uncertainty-laden systems, fostering both the accuracy and reliability of scientific and engineering procedures.

\subsection{Intuitive explanation of DVA mechanism}

Our DVA method is built upon the principle of the sample variance estimation \citep{degroot2012probability} describing that for a large number of i.i.d. random variables from a distribution with finite variance, the sample variance converges to the true variance of the distribution as the sample size increases. For instance, in the scenario of the zero mean noise case $(\sigma_{\epsilon}^2)$ with the sampled dataset $\tilde{D}=\{(x_i,y_i+\epsilon_i)\}_{i=1}^M$, the mean and variance of normalized noise $\{\epsilon_i/\sigma_{\epsilon}\}_{i=1}^M$ should converge to 0 and 1 when $M\rightarrow\infty$. This illustrates the concept that the larger the dataset becomes, the closer the normalized noise's average and variance align with these specific values.

Keeping that in mind, we parameterize the normalized noises (\{$\hat{\epsilon}_i\}$) and the variance $(\hat{\sigma}_{\epsilon}(\cdot))$ separately, where the former can be globally normalized to capture the random realization, and the latter accounts for the noise magnitude. After training, the learned variance function can be used for characterizing the noise dependence on $x$. By introducing the denoised label $(\Tilde{y}_i-\hat{\epsilon}_i\hat{\sigma}_{\epsilon}(x_i))$ into the DVA loss function in \Cref{eqn:denoisevarianceatt}, we aim to find a set of inferred noise that maximizes the likelihood of the denoised data matching the prediction. However, since the loss function involves the prediction model, the accuracy of the noise recovery depends on the prediction model's precision. As discussed in \Cref{sec:proposedmethod}, our approach yields a biased estimator but brings us ``one step closer'' to the true aleatoric uncertainty compared to the state-of-the-art VA method.

One might argue that the separate parameterization of the normalized noises and variance is over-parameterized, as we can directly model the unnormalized noise. Such approach is trivial because the denoised label will easily converge to the prediction assuming full flexibility of the variance. The neural network parameterization inherently regulates the smoothness of the variance function \cite{zi22smoothness} and limited a finite number of limited linear regions \cite{pmlr-v130-takai21a}, preventing trivial solutions and encouraging the learned variance to represent the true uncertainties.

\subsection{Limitations}
Despite the benefits, our approach has certain limitations. First, we introduce additional variables, requiring more memory and computational cost\footnote{The computation time only slightly increases compared with the VA method as most of the back-propagation paths are shared. For the age prediction task, the average training time per epoch is $34.7s$ for the VA method and $34.9s$ for the DVA method.}. Second, DVA assumes the zero mean of the sampled noise with the law of large numbers which requires a large sample size. When addressing heteroscedastic noise, we employ a segmented normalization technique to minimize the impact of data-dependent variance on the estimated normalized noise. Alternatively, the noise parameterization should be matched to the intrinsic heteroscedasticity of the system. In addition, this paper primarily concentrates on regression tasks, and the adaptation of these methods to classification tasks is beyond the scope of this work. 

\section{Conclusion}

In this study, we point out that the variance attenuation method widely used in deep learning, overestimates the aleatoric uncertainty. We propose a novel approach for estimating aleatoric uncertainty from data by considering the inferred noise and variance separately. Although our method still represents a biased estimation, our theoretical analysis demonstrates that it offers an improvement over the prior method in capturing the true aleatoric uncertainty, and we validate the improvement by quantitative experiments. In conclusion, our work presents a valuable contribution to the field of uncertainty estimation for deep neural networks and has the potential to serve as a standard procedure in aleatoric uncertainty quantification for machine learning tasks.

\clearpage
\section*{Acknowledgements}
This research is based upon the work supported by MIT-IBM Watson AI Lab. The authors thank Ching-Yun Irene Ko and Nima Dehmamy for the helpful discussions.
\bibliography{aaai24}

\begin{thebibliography}{26}
\providecommand{\natexlab}[1]{#1}

\bibitem[{Agustsson et~al.(2017)Agustsson, Timofte, Escalera, Baro, Guyon, and
  Rothe}]{agustsson2017apparent}
Agustsson, E.; Timofte, R.; Escalera, S.; Baro, X.; Guyon, I.; and Rothe, R.
  2017.
\newblock Apparent and real age estimation in still images with deep residual
  regressors on appa-real database.
\newblock In \emph{2017 12th IEEE International Conference on Automatic Face \&
  Gesture Recognition (FG 2017)}, 87--94. IEEE.

\bibitem[{Amini et~al.(2020)Amini, Schwarting, Soleimany, and
  Rus}]{amini2020deep}
Amini, A.; Schwarting, W.; Soleimany, A.; and Rus, D. 2020.
\newblock Deep evidential regression.
\newblock \emph{Advances in Neural Information Processing Systems}, 33:
  14927--14937.

\bibitem[{Blundell et~al.(2015)Blundell, Cornebise, Kavukcuoglu, and
  Wierstra}]{blundell2015weight}
Blundell, C.; Cornebise, J.; Kavukcuoglu, K.; and Wierstra, D. 2015.
\newblock Weight uncertainty in neural network.
\newblock In \emph{International conference on machine learning}, 1613--1622.
  PMLR.

\bibitem[{Chen et~al.(2018)Chen, Rubanova, Bettencourt, and
  Duvenaud}]{chen2018neuralode}
Chen, R.~T.; Rubanova, Y.; Bettencourt, J.; and Duvenaud, D.~K. 2018.
\newblock Neural ordinary differential equations.
\newblock \emph{Advances in neural information processing systems}, 31.

\bibitem[{Choi et~al.(2021)Choi, Elezi, Lee, Farabet, and
  Alvarez}]{choi2021active}
Choi, J.; Elezi, I.; Lee, H.-J.; Farabet, C.; and Alvarez, J.~M. 2021.
\newblock Active learning for deep object detection via probabilistic modeling.
\newblock In \emph{Proceedings of the IEEE/CVF International Conference on
  Computer Vision}, 10264--10273.

\bibitem[{DeGroot and Schervish(2012)}]{degroot2012probability}
DeGroot, M.~H.; and Schervish, M.~J. 2012.
\newblock \emph{Probability and statistics}.
\newblock Pearson Education.

\bibitem[{Der~Kiureghian and Ditlevsen(2009)}]{der2009aleatory}
Der~Kiureghian, A.; and Ditlevsen, O. 2009.
\newblock Aleatory or epistemic? Does it matter?
\newblock \emph{Structural safety}, 31(2): 105--112.

\bibitem[{Gal and Ghahramani(2016)}]{gal2016dropout}
Gal, Y.; and Ghahramani, Z. 2016.
\newblock Dropout as a bayesian approximation: Representing model uncertainty
  in deep learning.
\newblock In \emph{international conference on machine learning}, 1050--1059.
  PMLR.

\bibitem[{Gal et~al.(2016)}]{gal2016uncertaintythesis}
Gal, Y.; et~al. 2016.
\newblock Uncertainty in deep learning.

\bibitem[{Gelfand, Silverman et~al.(2000)}]{gelfand2000calculus}
Gelfand, I.~M.; Silverman, R.~A.; et~al. 2000.
\newblock \emph{Calculus of variations}.
\newblock Courier Corporation.

\bibitem[{Hern{\'a}ndez-Lobato and Adams(2015)}]{hernandez2015probabilistic}
Hern{\'a}ndez-Lobato, J.~M.; and Adams, R. 2015.
\newblock Probabilistic backpropagation for scalable learning of bayesian
  neural networks.
\newblock In \emph{International conference on machine learning}, 1861--1869.
  PMLR.

\bibitem[{H{\"u}llermeier and Waegeman(2021)}]{hullermeier2021aleatoric}
H{\"u}llermeier, E.; and Waegeman, W. 2021.
\newblock Aleatoric and epistemic uncertainty in machine learning: An
  introduction to concepts and methods.
\newblock \emph{Machine Learning}, 110: 457--506.

\bibitem[{Kurakin, Goodfellow, and Bengio(2018)}]{kurakin2018adversarial}
Kurakin, A.; Goodfellow, I.~J.; and Bengio, S. 2018.
\newblock Adversarial examples in the physical world.
\newblock In \emph{Artificial intelligence safety and security}, 99--112.
  Chapman and Hall/CRC.

\bibitem[{Lakshminarayanan, Pritzel, and
  Blundell(2017)}]{lakshminarayanan2017simple}
Lakshminarayanan, B.; Pritzel, A.; and Blundell, C. 2017.
\newblock Simple and scalable predictive uncertainty estimation using deep
  ensembles.
\newblock \emph{Advances in neural information processing systems}, 30.

\bibitem[{Liberzon(2011)}]{liberzon2011calculus}
Liberzon, D. 2011.
\newblock \emph{Calculus of variations and optimal control theory: a concise
  introduction}.
\newblock Princeton university press.

\bibitem[{Madry et~al.(2017)Madry, Makelov, Schmidt, Tsipras, and
  Vladu}]{madry2017towards}
Madry, A.; Makelov, A.; Schmidt, L.; Tsipras, D.; and Vladu, A. 2017.
\newblock Towards deep learning models resistant to adversarial attacks.
\newblock \emph{arXiv preprint arXiv:1706.06083}.

\bibitem[{Nix and Weigend(1994)}]{nix1994variance}
Nix, D.; and Weigend, A. 1994.
\newblock Estimating the mean and variance of the target probability
  distribution.
\newblock In \emph{Proceedings of 1994 IEEE International Conference on Neural
  Networks (ICNN'94)}, volume~1, 55--60 vol.1.

\bibitem[{Parker(2013)}]{parker2013ensemble}
Parker, W.~S. 2013.
\newblock Ensemble modeling, uncertainty and robust predictions.
\newblock \emph{Wiley Interdisciplinary Reviews: Climate Change}, 4(3):
  213--223.

\bibitem[{Seeb{\"o}ck et~al.(2019)Seeb{\"o}ck, Orlando, Schlegl, Waldstein,
  Bogunovi{\'c}, Klimscha, Langs, and Schmidt-Erfurth}]{seebock2019exploiting}
Seeb{\"o}ck, P.; Orlando, J.~I.; Schlegl, T.; Waldstein, S.~M.; Bogunovi{\'c},
  H.; Klimscha, S.; Langs, G.; and Schmidt-Erfurth, U. 2019.
\newblock Exploiting epistemic uncertainty of anatomy segmentation for anomaly
  detection in retinal OCT.
\newblock \emph{IEEE transactions on medical imaging}, 39(1): 87--98.

\bibitem[{Seitzer et~al.(2022)Seitzer, Tavakoli, Antic, and
  Martius}]{seitzer2022pitfalls}
Seitzer, M.; Tavakoli, A.; Antic, D.; and Martius, G. 2022.
\newblock On the pitfalls of heteroscedastic uncertainty estimation with
  probabilistic neural networks.
\newblock \emph{arXiv preprint arXiv:2203.09168}.

\bibitem[{Silberman et~al.(2012)Silberman, Hoiem, Kohli, and
  Fergus}]{silberman2012indoor}
Silberman, N.; Hoiem, D.; Kohli, P.; and Fergus, R. 2012.
\newblock Indoor segmentation and support inference from rgbd images.
\newblock \emph{ECCV (5)}, 7576: 746--760.

\bibitem[{Stirn et~al.(2023)Stirn, Wessels, Schertzer, Pereira, Sanjana, and
  Knowles}]{stirn23Faithful}
Stirn, A.; Wessels, H.; Schertzer, M.; Pereira, L.; Sanjana, N.; and Knowles,
  D. 2023.
\newblock Faithful Heteroscedastic Regression with Neural Networks.
\newblock In \emph{Proceedings of The 26th International Conference on
  Artificial Intelligence and Statistics}.

\bibitem[{Takai, Sannai, and Cordonnier(2021)}]{pmlr-v130-takai21a}
Takai, Y.; Sannai, A.; and Cordonnier, M. 2021.
\newblock On the number of linear functions composing deep neural network:
  Towards a refined definition of neural networks complexity.
\newblock In \emph{Proceedings of The 24th International Conference on
  Artificial Intelligence and Statistics}.

\bibitem[{Valdenegro-Toro and Mori(2022)}]{valdenegro2022deeper}
Valdenegro-Toro, M.; and Mori, D.~S. 2022.
\newblock A deeper look into aleatoric and epistemic uncertainty
  disentanglement.
\newblock In \emph{2022 IEEE/CVF Conference on Computer Vision and Pattern
  Recognition Workshops (CVPRW)}, 1508--1516. IEEE.

\bibitem[{Wang, Prakriya, and Jha(2022)}]{zi22smoothness}
Wang, Z.; Prakriya, G.; and Jha, S. 2022.
\newblock A Quantitative Geometric Approach to Neural-Network Smoothness.
\newblock In \emph{Advances in Neural Information Processing Systems}.

\bibitem[{Zhao et~al.(2023)Zhao, Rao, Liu, Liu, Zhou, and
  Lu}]{zhao2023unleashing}
Zhao, W.; Rao, Y.; Liu, Z.; Liu, B.; Zhou, J.; and Lu, J. 2023.
\newblock Unleashing Text-to-Image Diffusion Models for Visual Perception.
\newblock \emph{arXiv preprint arXiv:2303.02153}.

\end{thebibliography}
\clearpage
\onecolumn
\appendix
\section*{Appendix}
\section{Derivation of necessary conditions for local extremum}\label{appendix:theory}

\subsection{Derivation of extremum conditions for heteroscedastic variance attenuation label noise estimator}\label{appendix:likelihoodattvariational}

Consider an integral version of \Cref{eqn:loglikehoodatt}, where $p(x)$ denotes the probability distribution of $x$, and $\tilde{y}(x)$ is a random variable with a mean of $\mathbb{E}[\tilde{y}]=y=g(x)$ and variance $\mathbb{E}[(\tilde{y}-y)^2]=\sigma_\epsilon^2(x)$. 

\begin{equation}\label{eqn:loglikehoodattintegral}
        L_{va,int} =  \int_{\mathcal{X}} p(x)[\frac{(\tilde{y}(x)-\mu_{\theta_1}(x))^2}{2\hat{\sigma}^2_{\theta_2}(x)} + \frac{log(\hat{\sigma}^2_{\theta_2}(x))}{2} ]dx
\end{equation}

Fix the regression function $f_{\theta_1}(\cdot)$ and thus its mean prediction $\mu_{\theta_1}(\cdot)$. Let a continuous $h:\mathcal{X}\rightarrow \mathbb{R}$ be the increment of $\hat{\sigma}^2_{\theta_2}(\cdot)$ and $L_{va,int}(\hat{\sigma}^2_{\theta_2})$ be a functional of $\hat{\sigma}^2_{\theta_2}(\cdot)$. By the necessary condition for an extremum \citep{gelfand2000calculus}, when the differential functional $L_{va,int}[\hat{\sigma}^2_{\theta_2}]$ has an extremum, for all admissible $h$, we have

\begin{align}
   \delta L_{va,int}[\hat{\sigma}^2_{\theta_2}] =  \lim_{\epsilon\rightarrow 0}  \frac{L_{va,int}(\hat{\sigma}^2_{\theta_2}+\epsilon h)-L_{va,int}(\hat{\sigma}^2_{\theta_2})}{\epsilon} = 0
\end{align}

Calculate the derivative:
\begin{align}\label{eqn:variantionalderivative}
    \frac{d}{d\epsilon} L_{va,int}(\hat{\sigma}^2_{\theta_2}+\epsilon h) &=  \int_{\mathcal{X}} p(x)[\frac{-(\tilde{y}(x)-\mu_{\theta_1}(x))^2h}{2\hat{\sigma}^4_{\theta_2}(x)} + \frac{h}{2\hat{\sigma}^2_{\theta_2}(x)} ]dx  
\end{align}

\Cref{eqn:variantionalderivative} being zero for all admissible $h$ indicates that:
\begin{align}\label{eqn:flawedestimatorexpectheter}
    \hat{\sigma}^2_{\theta_2}(x) & = (\tilde{y}(x)-\mu_{\theta_1}(x))^2 \nonumber \\ 
    \implies \mathbb{E}[\hat{\sigma}^2_{\theta_2}(x)]&=\mathbb{E}[ (\tilde{y}(x)-\mu_{\theta_1}(x))^2] \nonumber \\ 
    & = \mathbb{E}_{\epsilon(x),f_{\theta_1}(x)}[(g(x)+\epsilon(x)-\mu_{\theta_1}(x))^2] \nonumber \\ 
    & = (g(x)-\mu_{\theta_1}(x))^2+\sigma_{\epsilon}^2(x)
\end{align}
where we slightly abuse the notation and let $\epsilon(x)$ be random data noise at $x$. $\sigma_{\epsilon}^2(x)$ is the variance of the data noise, $\mu_{\theta_1}(x)$ is the variance of the epistemic regression model.

\Cref{eqn:flawedestimatorexpectheter} comes to a similar conclusion 
to \Cref{eqn:flawedestimatorexpecthomo}.

\subsection{Derivation of necessary conditions for homoscedastic DVA label noise estimator}\label{appendix:KKTderivation}

We rewrite \Cref{eqn:kktcondition} with circled equation numbering:

\begin{align}
     \frac{-2(\Tilde{y}_i-\hat{\epsilon}_i\hat{\sigma}_{\epsilon,\theta_3} -\mu_{\theta_1}(x_i))\hat{\sigma}_{\epsilon,\theta_3}}{2\hat{\sigma}^2_{\theta_2}}+\lambda_1+2\lambda_2 \hat{\epsilon}_i&=0, \text{ for } i = 1,2...M \cdots\cdots\textcircled{1} \nonumber \\
        \Sigma_{i=1}^M \frac{-2(\Tilde{y}_i-\hat{\epsilon}_i\hat{\sigma}_{\epsilon,\theta_3} -\mu_{\theta_1}(x_i))\hat{\epsilon}_i}{2\hat{\sigma}^2_{\theta_2}}&=0 \cdots\cdots\textcircled{2} \nonumber  \\
        \Sigma_{i=1}^M \frac{(\Tilde{y}_i-\hat{\epsilon}_i\hat{\sigma}_{\epsilon,\theta_3} -\mu_{\theta_1}(x_i))^2}{M} & = \hat{\sigma}^2_{\theta_2} \cdots\cdots\textcircled{3}
\end{align}

Summing up $\textcircled{1}$, we have:

\begin{align*}
    \Sigma_{i=1}^M [\frac{-2(\Tilde{y}_i-\hat{\epsilon}_i\hat{\sigma}_{\epsilon,\theta_3} -\mu_{\theta_1}(x_i))\hat{\sigma}_{\epsilon,\theta_3}}{2\hat{\sigma}^2_{\theta_2}}+\lambda_1+2\lambda_2 \hat{\epsilon}_i] & = 0 \\
    \Sigma_{i=1}^M [\frac{-2(\Tilde{y}_i -\mu_{\theta_1}(x_i))\hat{\sigma}_{\epsilon,\theta_3}}{2\hat{\sigma}^2_{\theta_2}}]+\frac{-\hat{\sigma}_{\epsilon,\theta_3}^2}{\hat{\sigma}^2_{\theta_2}}\Sigma_{i=1}^M \hat{\epsilon}_i+M\lambda_1+2\lambda_2 \Sigma_{i=1}^M \hat{\epsilon}_i & = 0 \\
    \lambda_1 = \frac{1}{M}\frac{\hat{\sigma}_{\epsilon,\theta_3}}{\hat{\sigma}^2_{\theta_2}}\Sigma_{i=1}^M (\Tilde{y}_i -\mu_{\theta_1}(x_i)) \cdots\cdots\textcircled{4}
\end{align*}

Multiply $\textcircled{1}$ by $\hat{\epsilon}_i$ and sum up together:

\begin{align*}
    \Sigma_{i=1}^M [\frac{-2(\Tilde{y}_i-\hat{\epsilon}_i\hat{\sigma}_{\epsilon,\theta_3} -\mu_{\theta_1}(x_i))\hat{\sigma}_{\epsilon,\theta_3}\hat{\epsilon}_i}{2\hat{\sigma}^2_{\theta_2}}+\lambda_1\hat{\epsilon}_i+2\lambda_2 \hat{\epsilon}_i^2] & = 0 \\
     \Sigma_{i=1}^M [\frac{-2(\Tilde{y}_i -\mu_{\theta_1}(x_i))\hat{\sigma}_{\epsilon,\theta_3}\hat{\epsilon}_i}{2\hat{\sigma}^2_{\theta_2}}]+\frac{\hat{\sigma}_{\epsilon,\theta_3}^2}{\hat{\sigma}^2_{\theta_2}}\Sigma_{i=1}^M \hat{\epsilon}_i^2+\lambda_1\Sigma_{i=1}^M \hat{\epsilon}_i +2\lambda_2 \Sigma_{i=1}^M \hat{\epsilon}_i^2 & = 0 \\
      \Sigma_{i=1}^M [\frac{-2(\Tilde{y}_i -\mu_{\theta_1}(x_i))\hat{\sigma}_{\epsilon,\theta_3}\hat{\epsilon}_i}{2\hat{\sigma}^2_{\theta_2}}]+\frac{M\hat{\sigma}_{\epsilon,\theta_3}^2}{\hat{\sigma}^2_{\theta_2}} +2\lambda_2M  & = 0 \cdots\cdots\textcircled{5}\\
\end{align*}

By $\textcircled{2}$ from \Cref{eqn:kktcondition} we have:

\begin{align*}
     \Sigma_{i=1}^M (\Tilde{y}_i-\hat{\epsilon}_i\hat{\sigma}_{\epsilon,\theta_3} -\mu_{\theta_1}(x_i))\hat{\epsilon}_i &= 0\\
     \Sigma_{i=1}^M (\Tilde{y}_i -\mu_{\theta_1}(x_i))\hat{\epsilon}_i &= \hat{\sigma}_{\epsilon,\theta_3} \Sigma_{i=1}^M \hat{\epsilon}_i^2 = M\hat{\sigma}_{\epsilon,\theta_3} \cdots\cdots\textcircled{6}
\end{align*}

Substitute $\textcircled{6}$ to $\textcircled{5}$ we have:
\begin{align*}
    \lambda_2 & = 0 \cdots\cdots\textcircled{7}
\end{align*}

Substitute $\textcircled{4},\textcircled{7}$ to $\textcircled{1}$ we have:
\begin{align*}
    \frac{-2(\Tilde{y}_i-\hat{\epsilon}_i\hat{\sigma}_{\epsilon,\theta_3} -\mu_{\theta_1}(x_i))\hat{\sigma}_{\epsilon,\theta_3}}{2\hat{\sigma}^2_{\theta_2}}+\lambda_1&=0\\
    \frac{(\Tilde{y}_i-\hat{\epsilon}_i\hat{\sigma}_{\epsilon,\theta_3} -\mu_{\theta_1}(x_i))\hat{\sigma}_{\epsilon,\theta_3}}{\hat{\sigma}^2_{\theta_2}} & = \frac{1}{M}\frac{\hat{\sigma}_{\epsilon,\theta_3}}{\hat{\sigma}^2_{\theta_2}}\Sigma_{i=1}^M (\Tilde{y}_i -\mu_{\theta_1}(x_i))\\
    \frac{1}{\hat{\sigma}_{\epsilon,\theta_3}} [\Tilde{y}_i-\mu_{\theta_1}(x_i)-\frac{1}{M}\Sigma_{i=1}^M (\Tilde{y}_i -\mu_{\theta_1}(x_i))] & =\hat{\epsilon}_i  \cdots\cdots\textcircled{8}
\end{align*}
By $\textcircled{8}$, we have 

\begin{align*}
    \hat{\epsilon}_i \hat{\sigma}_{\epsilon,\theta_3}& =\Tilde{y}_i-\mu_{\theta_1}(x_i)-\frac{1}{M}\Sigma_{i=1}^M (\Tilde{y}_i -\mu_{\theta_1}(x_i)) 
\end{align*}

Substitute $\textcircled{8}$ to $\textcircled{5}$ we have:
\begin{align*}
    \Sigma_{i=1}^M \{(\Tilde{y}_i -\mu_{\theta_1}(x_i))\frac{1}{\hat{\sigma}_{\epsilon,\theta_3}} [\Tilde{y}_i-\mu_{\theta_1}(x_i)-\frac{1}{M}\Sigma_{i=1}^M (\Tilde{y}_i -\mu_{\theta_1}(x_i))] \} &= M\hat{\sigma}_{\epsilon,\theta_3} \\
    \Sigma_{i=1}^M (\Tilde{y}_i -\mu_{\theta_1}(x_i))^2-\frac{1}{M}[\Sigma_{i=1}^M (\Tilde{y}_i -\mu_{\theta_1}(x_i))]^2&= M\hat{\sigma}_{\epsilon,\theta_3}^2 \\
    \frac{1}{M}\Sigma_{i=1}^M (\Tilde{y}_i -\mu_{\theta_1}(x_i))^2-\frac{1}{M^2}[\Sigma_{i=1}^M (\Tilde{y}_i -\mu_{\theta_1}(x_i))]^2&= \hat{\sigma}_{\epsilon,\theta_3}^2 \cdots\cdots\textcircled{9}
\end{align*}

Let us derive two useful equalities:
\begin{align*}
    & \mathbb{E}[\frac{1}{M}\Sigma_{i=1}^M (\Tilde{y}_i -\mu_{\theta_1}(x_i))^2]\\
    & = \mathbb{E}[\frac{1}{M}\Sigma_{i=1}^M (y_i+\epsilon_i -\mu_{\theta_1}(x_i))^2]\\
    & = \sigma_\epsilon^2+\mathbb{E}[(y-\mu_{\theta_1}(x))^2] \cdots\cdots\textcircled{10}\\
    & \mathbb{E}[\frac{1}{M^2}[\Sigma_{i=1}^M (\Tilde{y}_i -\mu_{\theta_1}(x_i))]^2] \\
    & = \mathbb{E}[\frac{1}{M^2}[\Sigma_{i=1}^M (y_i+\epsilon_i -\mu_{\theta_1}(x_i))]^2]\\
        & = \mathbb{E}[\frac{1}{M^2}\Sigma_{i=1}^M \Sigma_{j=1}^M (y_i+\epsilon_i -\mu_{\theta_1}(x_i))(y_j+\epsilon_j -\mu_{\theta_1}(x_j))]\\
    & = \frac{1}{M^2}\mathbb{E}[\Sigma_{i=1}^M \Sigma_{j\neq i} (y_i+\epsilon_i -\mu_{\theta_1}(x_i))(y_j+\epsilon_j -\mu_{\theta_1}(x_j))+\Sigma_{i=1}^M (y_i+\epsilon_i -\mu_{\theta_1}(x_i))^2]\\
    & = \frac{1}{M^2} [M(M-1)\mathbb{E}^2[y-\mu_{\theta_1}(x)]+M\mathbb{E}[(y-\mu_{\theta_1}(x))^2+\sigma_\epsilon^2]\\
    & = \frac{M-1}{M}\mathbb{E}^2[y-\mu_{\theta_1}(x)]+\frac{1}{M}\mathbb{E}[(y-\mu_{\theta_1}(x))^2] +\frac{1}{M} \sigma_\epsilon^2 \cdots\cdots\textcircled{11}
\end{align*}

Substitute $\textcircled{10},\textcircled{11}$ into $\textcircled{9}$, taking expectation, we have 
\begin{align*}
    \mathbb{E}[\hat{\sigma}_{\epsilon,\theta_3}^2] & = \mathbb{E}[\frac{1}{M}\Sigma_{i=1}^M (\Tilde{y}_i -\mu_{\theta_1}(x_i))^2-\frac{1}{M^2}[\Sigma_{i=1}^M (\Tilde{y}_i -\mu_{\theta_1}(x_i))]^2]\\
    & = \sigma_\epsilon^2+\mathbb{E}[(y-\mu_{\theta_1}(x))^2]-   \frac{M-1}{M}\mathbb{E}^2[y-\mu_{\theta_1}(x)]-\frac{1}{M}\mathbb{E}[(y-\mu_{\theta_1}(x))^2] -\frac{1}{M} \sigma_\epsilon^2\\
    & = \frac{M-1}{M} \sigma_\epsilon^2+\frac{M-1}{M}\mathbb{E}[(y-\mu_{\theta_1}(x))^2]-\frac{M-1}{M}\mathbb{E}^2[y-\mu_{\theta_1}(x)]\\
    & = \frac{M-1}{M} \sigma_\epsilon^2+\frac{M-1}{M}\text{Var}[y-\mu_{\theta_1}(x))]
\end{align*}

Substitute $\textcircled{8}$ into $\textcircled{3}$ from \Cref{eqn:kktcondition}, we have:
\begin{align*}
    \hat{\sigma}^2_{\theta_2} & = \Sigma_{i=1}^M \frac{(\Tilde{y}_i-\hat{\epsilon}_i\hat{\sigma}_{\epsilon,\theta_3} -\mu_{\theta_1}(x_i))^2}{M} \\
    & = \Sigma_{i=1}^M \frac{(\Tilde{y}_i-[\Tilde{y}_i-\mu_{\theta_1}(x_i)-\frac{1}{M}\Sigma_{i=1}^M (\Tilde{y}_i -\mu_{\theta_1}(x_i))]  -\mu_{\theta_1}(x_i))^2}{M} \\
    & = \Sigma_{i=1}^M \frac{(-\frac{1}{M}\Sigma_{i=1}^M (\Tilde{y}_i -\mu_{\theta_1}(x_i)))^2}{M}\\
    & = (\frac{1}{M}\Sigma_{i=1}^M (\Tilde{y}_i -\mu_{\theta_1}(x_i)))^2
\end{align*}
Substitute $\textcircled{11}$ into the above equation and take expectation, we have 

\begin{align*}
    \mathbb{E}[\hat{\sigma}^2_{\theta_2}]     & = \frac{M-1}{M}\mathbb{E}^2[y-\mu_{\theta_1}(x)]+\frac{1}{M}\mathbb{E}[(y-\mu_{\theta_1}(x))^2] +\frac{1}{M} \sigma_\epsilon^2 \\
\end{align*}

\subsection{Justification of necessary conditions for heteromoscedastic DVA label noise estimator}\label{appendix:kktderivationvariational}

Similar to the treatment in \Cref{appendix:likelihoodattvariational}, consider the integral version of \Cref{eqn:denoisevarianceatt}.

\begin{align}\label{eqn:kktconditionvariational}
        L_{dva,int} & =  \int_{\mathcal{X}} p(x)[\frac{(\tilde{y}(x)-\hat{\epsilon}(x)\hat{\sigma}_{\epsilon,\theta_3}(x)\mu_{\theta_1}(x))^2}{2\hat{\sigma}^2_{\theta_2}(x)} + \frac{log(\hat{\sigma}^2_{\theta_2}(x))}{2} ]dx \nonumber \\
        & s.t. \text{ }\int_{ \mathcal{X}} \hat{\epsilon}(x)dx = 0;\int_{ \mathcal{X}} \hat{\epsilon}^2(x)=1
\end{align}

By the necessary condition of extreme points with integral constraints \citep{liberzon2011calculus},
\begin{align}\label{eqn:kktconditionheter}
     \frac{-2(\Tilde{y}(x)-\hat{\epsilon}(x)\hat{\sigma}_{\epsilon,\theta_3}(x) -\mu_{\theta_1}(x))\hat{\sigma}_{\epsilon,\theta_3}(x)}{2\hat{\sigma}^2_{\theta_2}(x)}+\lambda_1+2\lambda_2 \hat{\epsilon}(x)&=0, \nonumber \\
        \frac{-2(\Tilde{y}(x)-\hat{\epsilon}(x)\hat{\sigma}_{\epsilon,\theta_3}(x) -\mu_{\theta_1}(x))\hat{\epsilon}(x)}{2\hat{\sigma}^2_{\theta_2}(x)}&=0  \nonumber  \\
        (\Tilde{y}(x)-\hat{\epsilon}(x)\hat{\sigma}_{\epsilon,\theta_3}(x) -\mu_{\theta_1}(x))^2& = \hat{\sigma}^2_{\theta_2} (x)
\end{align}
This optimization is over-parameterized, as it doesn't explicitly require the continuity of $\hat{\sigma}{\epsilon,\theta_3}(x)$ and $\hat{\epsilon}(x)$. This observation is suggested by the second equality, which implies a possibility to adjust their values to achieve $(\Tilde{y}(x)-\hat{\epsilon}(x)\hat{\sigma}{\epsilon,\theta_3}(x) -\mu_{\theta_1}(x))=0$ for all $x$. However, it's important to note that this is a non-rigorous interpretation and should be treated cautiously. In practice, $\hat{\sigma}{\epsilon,\theta_3}(\cdot),\hat{\sigma}^2{\theta_2}(\cdot)$ are often modeled as continuous functions via neural networks. If we simplistically consider $\hat{\sigma}{\epsilon,\theta_3}(\cdot),\hat{\sigma}^2{\theta_2}(\cdot)$ as constants and adopt the assumption of a unit variance constraint for $\hat{\epsilon}(x)$ within a small finite interval around $x:[x-\delta,x+\delta]$, then the problem may superficially resemble a homoscedastic problem in the integral setup. This interpretation, while potentially useful as a heuristic, lacks a rigorous mathematical foundation and should be validated with further analytical or empirical work.
\begin{align}\label{eqn:kktconditionheter2}
     \frac{-2(\Tilde{y}(x)-\hat{\epsilon}(x)\hat{\sigma}_{\epsilon,\theta_3} -\mu_{\theta_1}(x))\hat{\sigma}_{\epsilon,\theta_3}}{2\hat{\sigma}^2_{\theta_2}}+\lambda_1+2\lambda_2 \hat{\epsilon}(x)&=0, \forall x\in[x-\delta,x+\delta]\nonumber \\
        \int_{[x-\delta,x+\delta]}\frac{-2(\Tilde{y}(x)-\hat{\epsilon}(x)\hat{\sigma}_{\epsilon,\theta_3}(x) -\mu_{\theta_1}(x))\hat{\epsilon}(x)}{2\hat{\sigma}^2_{\theta_2}}dx&=0  \nonumber  \\
        \int_{[x-\delta,x+\delta]}(\Tilde{y}(x)-\hat{\epsilon}(x)\hat{\sigma}_{\epsilon,\theta_3} -\mu_{\theta_1}(x))^2dx& = \hat{\sigma}^2_{\theta_2} 
\end{align}

This is the integral version of the global homoscedastic problem in \Cref{appendix:KKTderivation}. Following the similar derivation and taking the expectation over the random noises, we will have:

\begin{align}\label{eqn:heterdvavariance}
    \mathbb{E}[\hat{\sigma}_{\epsilon,\theta_3}^2] &  = \mathbb{E}_{[x-\delta,x+\delta]} [\sigma_\epsilon^2(x)]+\text{Var}_{[x-\delta,x+\delta]}[g(x)-\mu_{\theta_1}(x))]
\end{align}
where $g(x)$ is the true mapping function and $\mu_{\theta_1}(x)$ is the mean of the stochastic prediction model.

As the number of observations increases, the estimated variance $\hat{\sigma}_{\epsilon,\theta_3}^2$ approaches its expected value. The second term in \Cref{eqn:heterdvavariance} shrinks to zero when $\delta\rightarrow0$ as $g(x)-\mu_{\theta_1}(x)$ is also continuous with respect to $x$. In other words, \Cref{eqn:heterdvavariance} suggests that if we have a large number of observations within a small interval, the estimated variance $\hat{\sigma}_{\epsilon,\theta_3}^2$ converges to the true aleatoric uncertainty. This is an intuitive outcome that can be obtained by computing the variances of the sample labels. However, the vanilla variance attenuation estimator in \Cref{eqn:flawedestimatorexpectheter} is still flawed and cannot accurately identify the correct variance, as the bias term depends on the prediction model.

\newpage
\section{Algorithm details}\label{appendix:algo}
\RestyleAlgo{ruled}

\begin{algorithm}[hbt!]
\caption{Normalized gradient descent denoising for homoscedastic label noise with global noise normalization}\label{alg:gradient_ydenoising}

\SetKwInOut{Input}{Input}
\SetKwInOut{Output}{Output}
\Input{mini batch size $B$, learning rate $\eta$, stochastic prediction model with parameter $\theta_1$}
\KwData{$\{x_i,\tilde{y}_i\}$ for $i=1...M$}
initialize $\theta_2,\theta_3,\hat{\epsilon}_i $ for $i=1...M$\;

\For{$epoch =1,2...$}{
    $loss \leftarrow0 $\;
    generate a random permutation of $M$ indices $s_M$\;
    \For{$j =1,2...M/B$}{
        \For{$l =(j-1)B+1...jB$}{ 
        $k=s_M[l]$\;
        sample the mean of prediction $\hat{\mu}_{\theta_1}(x_{k})$ from prediction model\;
        $loss \leftarrow loss + \frac{(\Tilde{y}_k-\hat{\epsilon}_k\hat{\sigma}_{\theta_3,\epsilon}-\hat{\mu}_{\theta_1}(x_{k}))^2}{2\hat{\sigma}^2_{\theta_2}}+\frac{log(\hat{\sigma}^2_{\theta_2})}{2}$\;
        }
        $loss \leftarrow loss/B$\;
         $\theta_2 \leftarrow \theta_2-\eta \frac{dloss}{d\theta_2}$\;
         $\theta_3 \leftarrow \theta_3-\eta \frac{dloss}{d\theta_3}$\;
         $k=s_M[l],\hat{\epsilon}_k \leftarrow \hat{\epsilon}_k-\eta \frac{dloss}{d\hat{\epsilon}_k}, \text{ for } l=(j-1)B+1...jB$ \;
    }
        
        $\mu \leftarrow \sum_{i=1}^M \hat{\epsilon}_i/M; var \leftarrow \sum_{i=1}^M (\hat{\epsilon}_i-\mu)^2/M$ \;
        $\hat{\epsilon}_i \leftarrow \frac{\hat{\epsilon}_i-\mu}{\sqrt{var}}, \text{ for } i=1,2...M. $
}
\end{algorithm}

\begin{algorithm}[hbt!]\label{alg:gradient_ydenoising_segmentnorm}
\caption{Normalized gradient descent denoising for heteroscedastic label noise with segmented noise normalization}

\SetKwInOut{Input}{Input}
\SetKwInOut{Output}{Output}
\Input{mini batch size $B$, learning rate $\eta$, stochastic prediction model with parameter $\theta_1$, number of segments $G$}
\KwData{$\{x_i,\tilde{y}_i\}$ for $i=1...M$}
initialize $\theta_2,\theta_3,\hat{\epsilon}_i $ for $i=1...M$\;
rank all indices by $x$ value and segment the ranked indices into $G$ segments $\{h_1,h_2...h_g\}$\;
\For{$epoch =1,2...$}{
    $loss \leftarrow0 $\;
    generate a random permutation of $M$ indices $s_M$\;
    
    \For{$j =1,2...M/B$}{
        \For{$l =(j-1)B+1...jB$}{ 
        $k=s_M[l]$\;
        sample the mean of prediction $\hat{\mu}_{\theta_1}(x_{k})$ from prediction model\;
        $loss \leftarrow loss + \frac{(\Tilde{y}_k-\hat{\epsilon}_k\hat{\sigma}_{\theta_3,\epsilon}(x_k)-\hat{\mu}_{\theta_1}(x_{k}))^2}{2\hat{\sigma}^2_{\theta_2}(x_k)}+\frac{log(\hat{\sigma}^2_{\theta_2}(x_k))}{2}$\;
        }
        $loss \leftarrow loss/B$\;
         $\theta_2 \leftarrow \theta_2-\eta \frac{dloss}{d\theta_2}$\;
         $\theta_3 \leftarrow \theta_3-\eta \frac{dloss}{d\theta_3}$\;
         $k=s_M[l],\hat{\epsilon}_k \leftarrow \hat{\epsilon}_k-\eta \frac{dloss}{d\hat{\epsilon}_k}, \text{ for } l=(j-1)B+1...jB$\;
    }
        \For{$g =1,2...G$}{ 
        $\mu \leftarrow \sum_{i\in h_g} \hat{\epsilon}_i*G/M; var \leftarrow \sum_{i\in h_g}(\hat{\epsilon}_i-\mu)^2*G/M$\;
        $\hat{\epsilon}_i \leftarrow \frac{\hat{\epsilon}_i-\mu}{\sqrt{var}}, \text{ for } i\in h_g. $
        }
}
\end{algorithm}

\begin{algorithm}[hbt!]
\caption{Normalized gradient descent denoising for homoscedastic input noise}\label{alg:gradient_xdenoising}

\SetKwInOut{Input}{Input}
\SetKwInOut{Output}{Output}
\Input{mini batch size $B$, learning rate $\eta$, stochastic prediction model with parameter $\theta_1$ with sampler size $N$}
\KwData{$\{\tilde{x}_i,y_i\}$ for $i=1...M$}
initialize $\theta_3,\hat{\epsilon}_i $ for $i=1...M$\;

\For{$epoch =1,2...$}{
    $loss \leftarrow0 $
    generate a random permutation of $M$ indices $s_M$\;
    \For{$j =1,2...M/B$}{
        \For{$l =(j-1)B+1...jB$}{ 
        $k=s_M[l]$\;
        \For{$n =1,2...N$}{
        sample the weight of the stochastic prediction model or select a base learner from ensemble models, let the sampled deterministic model be $\hat{f}_{\theta_1}(\cdot)$\;
        $loss \leftarrow loss + (y_k-\hat{f}_{\theta_1}(\tilde{x}_{k}-\hat{\epsilon}_k\hat{\sigma}_{\theta_3,\epsilon}))^2$\;
        }
        }
        $loss \leftarrow loss/B/N$\;
         $\theta_3 \leftarrow \theta_3-\eta \frac{dloss}{d\theta_3}$\;
         $k=s_M[l],\hat{\epsilon}_k \leftarrow \hat{\epsilon}_k-\eta \frac{dloss}{d\hat{\epsilon}_k}, \text{ for } l=(j-1)B+1...jB$ \;
    }
        
        $\mu \leftarrow \sum_{i=1}^M \hat{\epsilon}_i/M; var \leftarrow \sum_{i=1}^M (\hat{\epsilon}_i-\mu)^2/M$ \;
        $\hat{\epsilon}_i \leftarrow \frac{\hat{\epsilon}_i-\mu}{\sqrt{var}}, \text{ for } i=1,2...M. $
}
\end{algorithm}

\newpage
\section{Experiment details}\label{appendix:experimentdetails}
\subsection{Toy example}
\quad
\textbullet Neural network structure:
All the prediction modeling and variance modeling neural networks mentioned in the experiments are fully connected neural networks, with 1 hidden layer of 100 neurons and tanh activations. For the ensemble models, we use 5 base networks. For the Bayesian neural networks, we use the standard scale mixture parameter prior and sampling setup in \cite{blundell2015weight}. The sampling number for the Bayesian neural network is also 5.

\textbullet Data preparation:
For the training dataset, we sample 1000 points for input $x$ from a uniform distribution $U[1,9]$ and calculate the corresponding label $y$. A zero mean Gaussian noise is added to either input or label depending on different configurations. The testing dataset is drawing from $U-[1,11]$ for 1000 points.

\textbullet Training schedule:
For both prediction modeling and variance modeling, we use a learning rate of $0.01$ and an Adam optimizer for 200 epochs across all the training procedures.

\textbullet Computational resource:
We use a single RTX 2080 Super graphic card for all the toy example computations.

\subsection{Dynamical system}
The neural network structure, training procedure and computational resources are identical to the toy example case. The integrator we used in the data generation and neuralODE application is Runge-Kutta (4,5) method.

\textbullet Data preparation: we sample 100 input initial points $x_0$ from uniform distribution $U[1,9]$ and simulate the dynamical system $\dot{x}=x(1+sin(x))$ from initial conditions for $5$ seconds. We then collect trajectory observation with 0.1 second intervals with Gaussian noise as training data. 

\subsection{Depth estimation}
\quad
\textbullet Neural network structure: 
We use Visual Perception with a pre-trained Diffusion (VPD) model \cite{zhao2023unleashing} trained on the NYU Depth v2 dataset as the prediction model. For the uncertainty estimation task, the aleatoric uncertainty is parameterized.

\textbullet Data preparation:
We use the NYU Depth v2 dataset \citep{silberman2012indoor}. It consists of 27k images. Each image in the dataset has a corresponding depth map, captured by a depth camera. These images are taken from a variety of indoor scenes, such as kitchens, bedrooms, and bathrooms. For uncertainty estimation, we add different levels of synthetic Gaussian noise to the label pixels and apply our DVA method to detect the noise level.

\textbullet Training schedule:
For variance modeling, we use a learning rate of 0.05 and an Adam optimizer for 30 epochs.

\textbullet Computational resource:
We use a single RTX 4090 graphic card for the age prediction tasks.

\subsection{Age prediction}
\quad
\textbullet Neural network structure: 
We use a ResNext-50 (32$\times$4d) pre-trained on ImageNet as a backbone model. During age prediction training, replace the last layer and fine-tune the last layer along with the backbone model on the APPA-REAL database. During uncertainty, we append an additional linear layer after the backbone model, which maps its outputs (with a dimension of 2048) to a logarithmic uncertainty measure. We then fine-tune only this linear layer.

\textbullet Data preparation: We use the APPA-REAL database \citep{agustsson2017apparent}. It consists of 7591 images. Each image in the database has been assigned an apparent age label based on an average of 38 individual votes per image.

\textbullet Training schedule:
For prediction modeling, we fine-tune the model for 100 epochs with an Adam optimizer of $lr=0.001$. For uncertainty modeling, we train the appended last layer for 20 epochs with an Adam optimizer of $lr=0.001$.

\textbullet Computational resource:
We use a single RTX 4090 graphic card for the age prediction tasks.

\section{Additional experiments on homoscedastic input noise}\label{appendix:inputnoise}

\begin{table}[htb!]
\centering
\caption{Aleatoric uncertainty estimation under homoscedastic input noise}\label{tab:xplusonebysinx_x}
\scalebox{0.8}{
\begin{tabular}{|l|*{6}{c|}}
        \hline
        \diagbox{Method}{Noise var. $(a^2)$}&0.5&1.0&2.0&4.0&8.0\\\hline
        DVA + Bayesian NN & 0.59$\pm$0.20  &  3.84$\pm$1.47  & 9.02$\pm$0.71  & 24.01$\pm$2.84 & 75.21$\pm$18.39 \\\hline
        DVA + ensemble&  0.88$\pm$0.18 & 1.59$\pm$0.35   & 5.57$\pm$1.49  & 11.68$\pm$5.14  &32.35$\pm$10.53 \\\hline
\end{tabular}}
\end{table}

We conducted a trial to extend the problem setting from label noise to homoscedastic input noise which the VA method is incapable to handle. We set the noise magnitude like: $\sigma_{\xi}(x)=a,\sigma_{\epsilon}(x)=0$. We adopt the denoising method to input noise, and the detailed algorithm is postponed to \Cref{appendix:algo}. Despite being capable of capturing the growing trend of noise level, the noise level is highly overestimated. The reason for this can be attributed to the characteristics of neural networks, where the denoising optimizer can move the inputs beyond the range of the data distribution with a similar prediction.

\end{document}